\documentclass[no-hf]{ceurart}

\usepackage{listings}
\usepackage{graphicx}
\usepackage{xcolor}
\usepackage{pgfplots}
\pgfplotsset{compat=1.17}
\begin{document}

\copyrightyear{2025}
\copyrightclause{Copyright for this paper by its authors.
  Use permitted under Creative Commons License Attribution 4.0
  International (CC BY 4.0).}

\conference{CLEF 2025 Working Notes, 9 -- 12 September 2025, Madrid, Spain}

\title{GemDetox at TextDetox CLEF 2025: \\ Enhancing a Massively Multilingual Model for Text Detoxification
on Low-resource Languages}

\title[mode=sub]{Notebook for the PAN Lab at CLEF 2025}

\author[1]{Trung Duc Anh Dang}[%
email=cls364@alumni.ku.dk,
url=https://github.com/ducanhdt/ducanhhbtt-detoxification2025/,
]

\author[1]{Ferdinando Pio D'Elia}[%
email=plh618@alumni.ku.dk,
]
\cormark[1]

\address[1]{Centre for Language Technology, University of Copenhagen, Denmark}

\cortext[1]{Corresponding author.}

\begin{abstract}
  As social-media platforms emerge and evolve faster than the regulations meant to oversee them, automated detoxification might serve as a timely tool for moderators to enforce safe discourse at scale. We here describe our submission to the PAN 2025 Multilingual Text Detoxification Challenge, which rewrites toxic single-sentence inputs into neutral paraphrases across 15 typologically diverse languages. Building on a 12B-parameter Gemma-3 multilingual transformer, we apply parameter-efficient LoRA SFT fine-tuning and prompting techniques like few-shot and Chain-of-Thought. Our multilingual training corpus combines 3 600 human-authored parallel pairs, 21 600 machine-translated synthetic pairs, and model-generated pairs filtered by Jaccard thresholds. At inference, inputs are enriched with three LaBSE-retrieved neighbors and explicit toxic-span annotations. Evaluated via Style Transfer Accuracy, LaBSE-based semantic preservation, and xCOMET fluency, our system ranks first on high-resource and low-resource languages. Ablations show +0.081 joint score increase from few-shot examples and +0.088 from basic CoT prompting. ANOVA analysis identifies language resource status as the strongest predictor of performance ($\eta^2$ = 0.667, p < 0.01). \\
  \textcolor{red}{Warning: This paper contains offensive and potentially triggering texts that only serve as illustrative examples.}
\end{abstract}


\maketitle

\section{Introduction}
The widespread use of digital communication and social media platforms has prompted the need, rather urgent, for moderation and content detoxification strategies that can be proved effective and easy to implement. Toxic language, hate speech, and harassment jeopardize safety and pluralism of online spaces, motivating the research community to fortify automated methods that could intervene at scale. The PAN 2025 Multilingual Detoxification Challenge \cite{bevendorff:2025}\cite{dementieva2025overview} offers a shared testbed for systems that aim to rephrase user posts into safer language while keeping their original meaning, and across multiple languages.

Large‑scale detoxification is not a purely technical exercise; it carries substantial social, political, and practical ramifications. In real-world scenarios, excessive detoxification interventions may be perceived as censorship, eroding trust and discouraging open dialogue; interventions may strip away the original message of its core purpose. The stakes are even higher for minorities and marginalized communities, who nowadays often rely on these platforms to voice political and social dissent and might consequently feel censored: as far as dissenting voices are concerned, detoxification could silence the very people it aims to protect. While we set these broader implications aside for the present study, they remain a key source of motivation for our work.

The Multilingual Text Detoxification task at PAN 2025 challenges participants to rewrite a toxic piece of text into a non‑toxic form while preserving as much of the original meaning as possible across 15 typologically diverse languages, ranging from English and Spanish to Tatar and Hinglish. Formally, the input is a single‑sentence text that contains at least one instance of toxic language, and the system must produce a semantically faithful paraphrase with neutral tone.
Notably, the competition constrains the notion of toxicity to explicit toxicity: obscene or offensive lexicon in which meaningful neutral content is still present. Implicit toxicity, such as sarcasm, coded hate speech, or passive‑aggressive formulations, is excluded\footnote{It warrants mention that such subtle toxicity is just as dangerous as a social phenomenon, and ample research has shown how ordinary terms and euphemisms can smuggle dehumanizing ideas into everyday speech, gradually normalizing hatred and exclusion. \cite{Pascale19} Yet, any problem is such only when we can carve it into a series of smaller and more manageable steps; this is our starting point.}.
We here present our approach, which builds directly on the 2024 paradigm\cite{dementieva2024} of fine-tuning large-scale multilingual pretrained transformer models for detoxification. We opted for a model with extensive multilingual pre-training yet lightweight features, and focused on maximizing the possibilities of Chain-of-Thought prompting as well as data augmentation.

The paper is organized as follows: in Section 2 we review related work; Section 3 describes our methodology; Section 4 details the experimental setup; Section 5 presents our automatic and qualitative results; and Section 6 discusses limitations and future directions. Our experiments show that, while our approach achieves strong performance across all languages, the gains from data augmentation are especially pronounced for the low-resource languages for which they were designed, narrowing their gap with high-resource counterparts.

\section{Related work}
Text detoxification, a specialized subfield of Text Style Transfer (TST), involves transforming toxic texts into neutral versions while maintaining semantic integrity and linguistic fluency. Initial research in detoxification was largely driven by prominent competitions such as the Jigsaw/Conversation AI Kaggle challenges (2018–2021) \cite{jigsaw2018toxiccomment}, providing substantial datasets and significantly advancing toxicity detection methods. Early models, such as \cite{nogueira-dos-santos-etal-2018-fighting}, employed unsupervised encoder-decoder architectures with cycle consistency losses for style transfer. Subsequent approaches, such as CondBERT and ParaGedi \cite{dale-etal-2021-text}, introduced unsupervised conditional masked language modeling and paraphrasing, setting benchmarks at the time. Further progress was marked by the ParaDetox corpus \cite{logacheva-etal-2022-paradetox}, which demonstrated that parallel data substantially improved the detoxification performance over purely unsupervised techniques.

Recently, transformer-based large language models (LLMs) have been largely utilized in detoxification research due to their powerful text generation capabilities. Notably, GPT-based models, including GPT-2, GPT-3, and GPT-4, demonstrated considerable efficacy when fine-tuned or used with few-shot prompting strategies \cite{pesaranghader2024gptdetoxincontextlearningbasedparaphraser}. Models such as GPT-DETOX have explored innovative in-context learning techniques including zero-shot, few-shot, and ensemble prompting, significantly outperforming earlier supervised and unsupervised methods.

Parallel corpus availability remains critical for enhancing detoxification methods. The creation of new datasets such as the multilingual MultiParaDetox corpus \cite{dementieva-etal-2024-multiparadetox} has significantly expanded detoxification research to include languages previously underrepresented, such as Hindi, Arabic, Chinese, and Amharic. Furthermore, sophisticated prompt-engineering frameworks, like CO-STAR \cite{teo2023costar}, have shown promising results by strategically guiding LLMs to enhance contextual and semantic coherence during detoxification. Recent advancements have also emphasized explainability and interpretability in detoxification processes. Very recently, a first automated explainable analysis across multiple languages was published \cite{dementieva-etal-2025-multilingual}, revealing common patterns and language-specific toxicity traits, and implemented Chain-of-Thought reasoning techniques to enhance the detoxification accuracy of LLMs.
Hybrid detoxification models combining editing-based and sequence-to-sequence approaches, as demonstrated in \cite{floto2023diffudetoxmixeddiffusionmodel} with DiffuDetox, leverage the complementary strengths of multiple techniques. These hybrid approaches have achieved state-of-the-art performance in automatic evaluations and showed good results  in human assessments compared to purely editing-based or sequence-to-sequence models.

Overall, text detoxification research now involves a broad spectrum of methodologies from unsupervised conditional generation and fine-tuned LLMs to explainable AI and advanced multilingual prompting strategies. Competitions such as the Multilingual Text Detoxification Task at PAN 2024 offer valuable opportunities to explore the capabilities and limitations of LLMs, particularly in their lightweight versions.

\section{Methods}
\subsection{Datasets}
\label{datasets-section}
This section details the construction of the training corpus we employed in all experiments.  The final corpus merges (i) the organiser‑provided parallel data, (ii) machine‑translated extensions for six missing languages, and (iii) synthetic pairs mined from a toxicity‑only collection.
The PAN~2025 Multilingual Text‑Detoxification shared task \cite{dementieva2025overview} supplies a parallel corpus (ParaDetox‑9).  It comprises 400 toxic–neutral sentence pairs for each of nine languages: English, Spanish, German, Chinese, Arabic, Hindi, Ukrainian, Russian and Amharic, for a total of 3\,600 pairs.  These data constitute the only human‑authored supervision available for the task.

\subsubsection{Data augmentation for unseen languages}
\label{sec:paradetoxmt6}
The remaining six target languages (Italian, French, Hebrew, Hinglish, Japanese and Tatar) are not covered by ParaDetox‑9. To extend coverage to these languages, we translated all the 3 600 original pairs (in the 9 different "seen" languages) with two publicly available neural machine‑translation systems: the RLM‑Hinglish Translator \cite{shah2022_rlm_hinglish_translator} for Hinglish, and the NLLB‑200 (3.3~B‑parameter) \cite{nllbteam2022nllb} for Italian, French, Hebrew, Japanese and Tatar.

In all cases, each original toxic–neutral pair \((s_1,s_2)\) from ParaDetox‑9 was translated sentence-by-sentence into \((t_1,t_2)\). The whole corpus of 3 600 source pairs was translated for each of the six unseen languages, thereby obtaining 21 600 pairs in total. This resulting corpus, named ParaDetox-MT6, is publicly released alongside our code and models \cite{anhdtd2025_paradetox_mt6}.

\subsubsection{Additional synthetic data generation}
\label{sec:toxicitysynth15}
We incorporated the Multilingual Toxicity Dataset \cite{textdetox2024_multilingual_toxicity}, which contains toxic sentences in 15 languages but no corresponding non‑toxic rewrites. For each toxic sentence we generated one or more neutralization candidates with our strongest detoxification model (see further, \ref{proposedmodel}) and retained only those pairs that satisfied the filtering criteria described below, \ref{sec:cleaning}. This procedure contributed additional supervision\footnote{It is worth noting that recent work has already proposed interesting and useful synthetic data generation pipelines for multilingual detoxification objectives; see \cite{moskovskiy2025synthdetoxmmodernllmsfewshot} for further details.}.

\subsubsection{Filtering and cleaning}
\label{sec:cleaning}
To ensure high‑quality supervision, all candidate pairs, both machine‑translated and model-generated, are processed by a uniform cleaning pipeline:
\begin{enumerate}
  \item Jaccard similarity filtering, where we compute the character‑level Jaccard index
  \(\displaystyle J(s_{\text{toxic}},s_{\text{clean}})\) over 5‑grams and discard pairs with $J\ge0.90$ to enforce lexical divergence.
  \item Removal duplication, where pairs whose toxic and clean sentences are identical (after Unicode canonicalisation and lower‑casing) are removed.
  \item Hinglish script filtering, where candidate Hinglish language pairs containing any Devanagari code point are eliminated to avoid mixed‑script noise\footnote {In social and conversational text, Hinglish is almost exclusively written in the Latin (Roman) alphabet, with English words and romanised Hindi freely mixed. Any Devanagari code points usually signal a switch to standard-script Hindi, introducing (i) vocabulary sparsity, because the Hinglish tokenizer is trained on Roman script, and (ii) label ambiguity by blurring the boundary between the Hinglish and Hindi tracks.}.
  \item Semantic preservation filter, where a quality check is introduced on the pipeline to ensure the quality of the candidate neutral rewrites. To achieve this, pairs from the machine-translated augmentation set are retained only if $J$$>$0.85, while pairs synthesised from toxicity-only collections must satisfy the slightly looser $J$$>$0.80 threshold.
\end{enumerate}

Lastly, to provide additional context during both training and evaluation, each toxic sentence is first embedded with LaBSE and associated with its three closest semantic neighbours in the same language.  The identifiers of those six neighbours are stored alongside the sentence, giving downstream models quick access to in-language and cross-language examples that convey a similar meaning.  A second enrichment pass applies a rule-based detector that extracts every explicit slur or profanity token and records them next to the sentence. The rule-based detector implements the same exact strategy of the Delete baseline provided for the PAN shared task.

\subsubsection{Corpus statistics}
\label{sec:stats}
Table~\ref{tab:corpus-stats} summarizes the number of retained pairs per language after cleaning.  Altogether, the final corpus comprises \mbox{$\approx$18\,000} toxic–neutral pairs covering all 15 languages included in the PAN shared task.

\begin{table}[t]
  \centering
  \scriptsize
  \setlength\tabcolsep{3pt} 
  \caption{Toxic–neutral sentence pairs before and after the cleaning pipeline.}
  \label{tab:corpus-stats}
  \begin{tabular}{lrr}
    \toprule
    \textbf{Language} & \textbf{Pairs before} & \textbf{Pairs after}\\
    \midrule
    English (en)   & 1 022 & 1 012\\
    Spanish (es)   & 1 363 & 1 362\\
    German (de)    & 1 185 & 1 142\\
    Chinese (zh)   &   770 &   770\\
    Arabic (ar)    & 1 022 & 1 019\\
    Hindi (hi)     &   568 &   568\\
    Ukrainian (uk) & 2 535 & 2 535\\
    Russian (ru)   & 1 565 & 1 520\\
    Amharic (am)   &   719 &   719\\
    Italian (it)   & 2 694 & 1 948\\
    French (fr)    & 2 375 & 1 666\\
    Hebrew (he)    & 1 772 & 1 037\\
    Hinglish (hin) & 1 135 &   561\\
    Japanese (ja)  &   983 &   661\\
    Tatar (tt)     & 1 951 & 1 275\\
    \midrule
    \textbf{Total} & \textbf{21 659} & \textbf{17 795}\\
    \bottomrule
  \end{tabular}
\end{table}

\subsection{Baselines}
We evaluate our method against the unsupervised baselines provided by the PAN shared task. The trivial Duplicate baseline simply echoes the toxic source sentence unchanged. The Delete baseline removes any term appearing in the multilingual toxic lexicon \cite{zeira2024_multilingual_toxic_lexicon} for each language  without further rewriting. The Backtranslation baseline performs a two-step cross–lingual transfer: the input is first translated into English using the distilled NLLB‑200 600M‑parameter model \cite{nllb2023_distilled_600m}, then detoxified with the English BART‑base‑detox \cite{snlp2023_bart_base_detox} checkpoint, and finally translated back into the original language via NLLB‑200 600M. Finally, zero-shot prompting baselines were also provided by the shared task committee: LLaMA‑3.1‑70B‑Instruct‑lorablated \cite{labonne2024_llama_instruct}, as well as OpenAI's GPT‑4‑0613, GPT‑4o‑2024‑08‑06, and o3‑mini‑2025‑01‑31 \cite{openai2025_gpt_models}.

\subsection{Proposed model}
\label{proposedmodel}
As our base model we adopt the Gemma-3-12B-Instruct architecture\footnote{In initial phases of our work, we tested other models as well: in quick prompt-only tests, the 7 B Qwen-2.5 and 27 B Gemma-3 variants detoxified more reliably than the 4 B editions, but the 12 B Gemma matched the 27 B’s quality while fitting on a single 24 GB card.
We therefore decided to adopt Gemma-3 especially because of its massively multilingual pre-training, which we believed could offer promising performance on zero-shot transferring. When presented to the public by the Google Dev team, Gemma3 was reported to be "available in 140 languages". Training data for the model was not made available by Google.} \cite{gemma3technicalreport}, quantised to 4-bit integer precision to cap memory usage below 24 GB while retaining the original 4 096-token rotary position scheme. At load time the Unsloth runtime converts activations to BF16 on Hopper-class GPUs and expands the context window to 4 028 tokens.
To enable parameter-efficient adaptation we insert low-rank adapters (rank = 16, scaling = 16, no dropout) into every language, attention and MLP sub-module, freezing the remaining weights. This makes only 0.55\% ($\approx$ 65M) of the 12B parameters trainable while preserving the expressiveness of the original network.

To better leverage the possibilities of the language model, we design a prompting strategy inspired by Chain-of-thought (CoT) principles, given the outstanding results these approaches have shown in recent years \cite{wei2023chainofthoughtpromptingelicitsreasoning}. Namely, the system message outlines a four-step detoxification instruction which prompts the model to (i) identify toxic element(s), (ii) retrieve the semantic content of the overall sentence, (iii) rewrite using neutral words, (iv) check that your output be non-toxic.
We created a base-prompt in English\footnote{The base prompt is available on our open-access repository on GitHub.}, and then passed it through OpenAI's o4‑mini‑high \cite{openai2025_gpt_models} on a sample of three languages in which we are proficient, either as native speakers or as second-language users: we manually checked them for consistency, and then proceeded to translate the base-prompt to the remaining languages using the same model. The result was 15 different prompts in the respective languages required by the PAN shared task.

Each training instance is therefore rendered as a three-turn process: the system message, the user supplying the toxic sentence with the language tag, the model's returned sentence pairs. 
For stronger supervision, the prompt prepends the semantically closest $k$ examples (three in our experiments) drawn from the same language, retrieved from the datasets described in \ref{datasets-section} by selecting the top three toxic sentences most similar in meaning, context, and phrasing to the target, yielding language-aware few-shot conditioning.
We format the model's output in a standardized JSON structure. This improves output consistency, facilitates reliable parsing, and simplifies downstream processing for evaluation and training data collection.

During optimization, tokens belonging to the system and user turns are masked so that the cross-entropy objective is evaluated solely on the assistant span, preventing leakage of latent reasoning. Training data are further filtered by high semantic overlap ($J$ $>$ 0.9), inequality between toxic and neutral sentences, and membership in a seven-language target set. After fine-tuning, we convert the model into the VLLM format for low-latency, high-throughput inference. At inference time, we sample three candidate rewrites per toxic input and, using the available reference neutral sentence, compute the joint score $J$ for each triple \(\langle\text{toxic input},\,\text{reference neutral},\,\text{model output}\rangle\); the candidate with the highest $J$-score is chosen as the final detoxified output.

\section{Experimental setup}
\subsection{Evaluation}
\label{original-evaluation}
For our experiments, we adhere to the evaluation protocol established by the shared task committee, which employs three primary metrics: Style Transfer Accuracy (STA), Content Preservation (SIM), and Fluency (FL)—each normalized to the interval [0,1].

Style Transfer Accuracy assesses both the absolute and relative non-toxicity of the generated output. Let $p_{\mathrm{gen}}$ denote the non-toxicity probability of the machine-generated sentence and $p_{\mathrm{ref}}^{(i)}$ for $i=1$ to $N$ the probabilities for each of the $N$ human-authored detoxifications, as computed by a fine-tuned XLM-RoBERTa binary classifier \cite{xlmr_toxicity_classifier}. We define:
\begin{equation}
\mathrm{STA} = \frac{p_{\mathrm{gen}} + \frac{1}{N} \sum_{i=1}^{N} \mathbb{I}\bigl(p_{\mathrm{gen}} \le p_{\mathrm{ref}}^{(i)}\bigr)}{2},
\end{equation}
where $\mathbb{I}(\cdot)$ is the indicator function. This formulation penalizes outputs that are more toxic than human references while preventing over-rewarding outputs that simply match reference non-toxicity.

Content Preservation quantifies semantic fidelity through a weighted cosine similarity of LaBSE embeddings \cite{labse}:
\begin{equation}
\mathrm{SIM} = 0.4 \times \mathrm{CosSim}\bigl(\mathrm{input}, \mathrm{output}_{\mathrm{gen}}\bigr) + 0.6 \times \mathrm{CosSim}\bigl(\mathrm{output}_{\mathrm{gold}}, \mathrm{output}_{\mathrm{gen}}\bigr).
\end{equation}
This balance ensures that the generated detoxification remains faithful both to the original toxic input and to the human-written reference. Lastly, Fluency is measured by the xCOMET\cite{xcomet_lite} metric $X_{\mathrm{comet\_fluency}}$, which in experiments conducted by the shared task organizers, has demonstrated near‑perfect correlation with human judgments of fluency in detoxified texts.

These metrics are combined into a single joint score $J$ per sample:
\begin{equation}
\resizebox{\textwidth}{!}{$
J = X_{\mathrm{comet\_fluency}}\bigl(\mathrm{input}, \mathrm{output}_{\mathrm{gold}}, \mathrm{output}_{\mathrm{gen}}\bigr) \times \bigl(0.4 \times \mathrm{CosSim}(\mathrm{input}, \mathrm{output}_{\mathrm{gen}}) + 0.6 \times \mathrm{CosSim}(\mathrm{output}_{\mathrm{gold}}, \mathrm{output}_{\mathrm{gen}})\bigr) \times \mathrm{STA}$}
\end{equation}

Here, the updated STA term incorporates human references by averaging the classifier’s non-toxicity probability on the generated output with its relative position among the reference scores. This refinement penalizes outputs that remain more toxic than human-authored detoxifications, while the weighted similarity term balances fidelity to both the original input and the human detoxification.

\subsection{Model training}
We conduct Low-Rank Adaptation (LoRA) fine-tuning \cite{hu2022lora} on our multilingual corpus (with a 900-pair validation set) in three iterative phases. LoRA fine-tunes a large backbone by freezing its original weights $W_{0}$ and learning a low-rank update:
\[
W = W_{0} + \tfrac{\alpha}{r}\,AB,\qquad
A \in \mathbb{R}^{d\times r},\;
B \in \mathbb{R}^{r\times k}.
\]

With $r=\alpha=16$ on the 12-B-parameter Gemma-3-12B backbone this adds $\approx$ 65 M trainable parameters (0.55\%).
Combined with 4-bit quantisation and Unsloth’s fused kernels, this keeps an 8-sample batch on a single NVIDIA A100 (80 GB) without gradient accumulation.
Adapters are inserted in every self-attention projection and the MLP gating/up/down projections; dropout is 0, bias is \textit{none}, seed = 3407.
We train in three iterative phases:

\begin{itemize}
\item Phase 1. We fine-tune across all 15 languages for 1 000 optimisation steps with 8-bit AdamW (peak $LR = 2 \times 10^{-5}$; 20 warm-up steps; linear decay; weight decay = 0.01; seed = 3 407). This broadly aligns the model across high-resource (seen) and low-resource (unseen) languages.

\item Phase 2. We apply a second fine-tuning pass using only data for the languages without available parallel data, running 1 000 steps at a reduced $LR = 5 \times 10^{-6}$ to avoid catastrophic interference, thereby preserving the first phase for high-resource languages. We save this adapter as Checkpoint II.

\item Phase 3. We apply the best model from Phase 2 to generate non-toxic rewrites for toxic-only sentences in the synthetic dataset (described above in \ref{sec:toxicitysynth15}), filtered by Jaccard $>$ 0.9, then fine-tune for a final 1 000 steps. We save this adapter as Checkpoint III.
\end{itemize}

At inference, languages without available parallel data are handled by the Checkpoint II, whereas languages without available parallel data are processed by the more specialized adapter from the third phase (Checkpoint III).

Across all phases (3 000 steps, \~3 hours total), memory is conserved via Unsloth’s gradient checkpointing and smart off-loading; checkpoints are saved every 100 steps, and token-level perplexity (training loss) is logged to Weights \& Biases at 20-step intervals. At inference, three candidate rewrites are sampled per input and the best one is chosen.

\section{Results}
Our final automatic results on the Test set are reported in Table \ref{tab:avgp_all} and Table \ref{tab:avgnp}, respectively highlighting the languages with human-annotated parallel data provided by the PAN shared task organizers, and those without.

Our model reaches its highest performance on German ($J$ = 0.798, 1st) and its lowest on Amharic ($J$ = 0.446, 8th) among the first batch of languages. The model also peaks on French ($J$ = 0.802, 1st) and falls to its minimum on Hinglish ($J$ = 0.511, 1st), a low‐resource language. Overall, as we will also outline below in Paragraph \ref{anova_paragraph}, the lowest results by far are reported in all of the low-resource languages.

\begin{table}[ht]
  \centering
  \caption{Automatic evaluation results across different model variants on languages with available parallel data from human annotators. The results refer to the final submission phase. In parentheses: rank per language.}
  \label{tab:avgp_all}
  \resizebox{\linewidth}{!}{%
    \begin{tabular}{lcccccccccc}
      \toprule
      Model                     & Avg J (rank) & en (rank) & es (rank) & de (rank) & zh (rank)
                                & ar (rank)  & hi (rank) & uk (rank) & ru (rank) & am (rank) \\
      \midrule
      Our submission            & 0.685 (1)  & 0.734 (3) & 0.686 (11) & 0.798 (1) & 0.618 (1)
                                & 0.718 (3)  & 0.619 (7) & 0.799 (2) & 0.749 (6) & 0.446 (10) \\
      baseline\_mt0             & 0.675 (5)  & 0.727 (6) & 0.696 (10) & 0.757 (6) & 0.543 (8)
                                & 0.715 (4)  & 0.627 (5) & 0.770 (6) & 0.754 (2) & 0.491 (1) \\
      baseline\_delete          & 0.536 (26) & 0.473 (29)& 0.603 (26)& 0.586 (25)& 0.516 (15)
                                & 0.611 (16) & 0.480 (25)& 0.581 (26)& 0.514 (28)& 0.461 (5) \\
      baseline\_gpt4            & 0.637 (12) & 0.708 (13)& 0.708 (5)& 0.728 (12)& 0.513 (16)
                                & 0.603 (17) & 0.605 (10)& 0.747 (11)& 0.706 (12)& 0.412 (16) \\
      baseline\_backtranslation & 0.481 (28) & 0.684 (18)& 0.528 (29)& 0.513 (29)& 0.290 (29)
                                & 0.438 (28) & 0.419 (28)& 0.498 (28)& 0.696 (13) & 0.265 (27) \\
      \bottomrule
    \end{tabular}%
  }
\end{table}

\begin{table}[ht]
  \caption{Automatic evaluation results across different model variants on languages with no available parallel data from human annotators. The results refer to the final submission phase. In parentheses: rank per language.}
  \label{tab:avgnp}
  \resizebox{\linewidth}{!}{%
      \begin{tabular}{lccccccc}
        \toprule
        Model                      & Avg J (rank) & it (rank) & ja (rank) & he (rank)
                                   & fr (rank)  & tt (rank) & hin (rank) \\
        \midrule
        Our submission            & 0.643 (1)  & 0.784 (1) & 0.674 (1) & 0.531 (1)
                                   & 0.802 (1)  & 0.556 (5) & 0.511 (1) \\
        baseline\_mt0             & 0.572 (12)  & 0.746 (8) & 0.582 (13) & 0.415 (23)
                                   & 0.760 (9)  & 0.580 (4) & 0.351 (18) \\
        baseline\_delete          & 0.510 (21) & 0.668 (21)& 0.441 (26)& 0.436 (18)
                                   & 0.518 (29) & 0.573 (7) & 0.425 (9) \\
        baseline\_gpt4            & 0.579 (9) & 0.742 (10)& 0.637 (7)& 0.513 (3)
                                   & 0.780 (6) & 0.468 (17)& 0.333 (19) \\
        baseline\_backtranslation & 0.342 (20) & 0.462 (29)& 0.241 (31)& 0.339 (28)
                                   & 0.626 (26) & 0.254 (30)& 0.133 (30) \\
        \bottomrule
      \end{tabular}%
  }
\end{table}

Interestingly, the baseline mT0 model \cite{dementieva2025overview} already achieves very competitive joint scores across most languages, ranking within the top 5 for 9 of the 17 tested languages and topping the leader-board on Amharic (0.491, 1st) despite zero fine-tuning. Its overall average $J_P$ of 0.675 places it only four places below our final system, underscoring the effectiveness of large, multilingual foundational models out of the box. GPT-4 also performs respectably (Avg $J_P$ 0.637, 12th), but falls behind MT0 on high-resource and cross-lingual pairs. These results highlight that simple, prompt-based baselines like MT0 remain strong contenders in detoxification tasks. For complete scores see Appendix~\ref{appendix:C}.

Moreover, Figure \ref{modelvariants} illustrates the progression of the joint score $J$ through our experiments and model development, through successive Gemma variants. The data refer to the Dev set simply for availability reasons, however, all data are perfectly comparable to our Test set as well.  Starting from the 4B baseline (0.562 / 0.514), we see that with
Few-shot prompting $J$ rises by +0.022 on parallel data and falls by –0.015 on no-parallel data (to 0.584 / 0.499). By scaling up to the 12B variant and using few-shot approach, we reach a further gain of +0.081 (parallel) and +0.071 (no-parallel), amounting 0.643 / 0.585. By implementing Chain-of-Thought, $J$ climbs to 0.650 / 0.592 ($\Delta$ +0.088 / +0.078 over baseline). During the two following phases (corresponding to the data augmentation phases) we register more modest improvements, ending at 0.685 / 0.607 and 0.692 / 0.642 respectively on the Dev set.

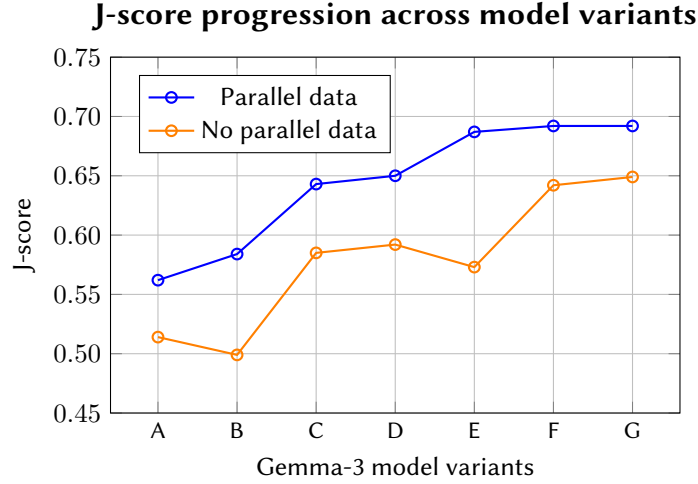
\begin{figure}[ht]
\centering
\begin{tikzpicture}
    \begin{axis}[
        width=0.57\textwidth,
        height=6.3cm,
        title={\large\textbf{{J-score progression across model variants}}},
        xlabel={Gemma-3 model variants},
        ylabel={J-score},
        xtick={0,1,2,3,4,5,6},
        xticklabels={A, B, C, D, E, F, G},
        xticklabel style={font=\footnotesize},
        ymin=0.45, ymax=0.75,
        ytick={0.45,0.50,0.55,0.60,0.65,0.70,0.75},
        scaled y ticks=false,
        yticklabel style={/pgf/number format/fixed, /pgf/number format/precision=2, /pgf/number format/zerofill},
        grid=both,
        legend style={at={(0.05,0.95)},anchor=north west}
    ]

    \addplot[color=blue,mark=o,thick] coordinates {
        (0,0.562) (1,0.584) (2,0.643) (3,0.650) (4,0.687) (5,0.692) (6,0.692)
    };
    \addlegendentry{Parallel data}

    \addplot[color=orange,mark=o,thick] coordinates {
        (0,0.514) (1,0.499) (2,0.585) (3,0.592) (4,0.573) (5,0.642) (6,0.649)
    };
    \addlegendentry{No parallel data}

    \end{axis}
\end{tikzpicture}
\caption{Model performance across phases: A = Gemma 4B zero-shot, B = Gemma 4B few-shot, C = Gemma 12B few-shot, D= Gemma 12B few-shot (additional phase), E = Gemma 12B CoT and few-shot, F = Gemma 12B CoT and few-shot (additional phase), G = Gemma 12B CoT and few-shot (second additional phase).}
\label{modelvariants}
\end{figure}

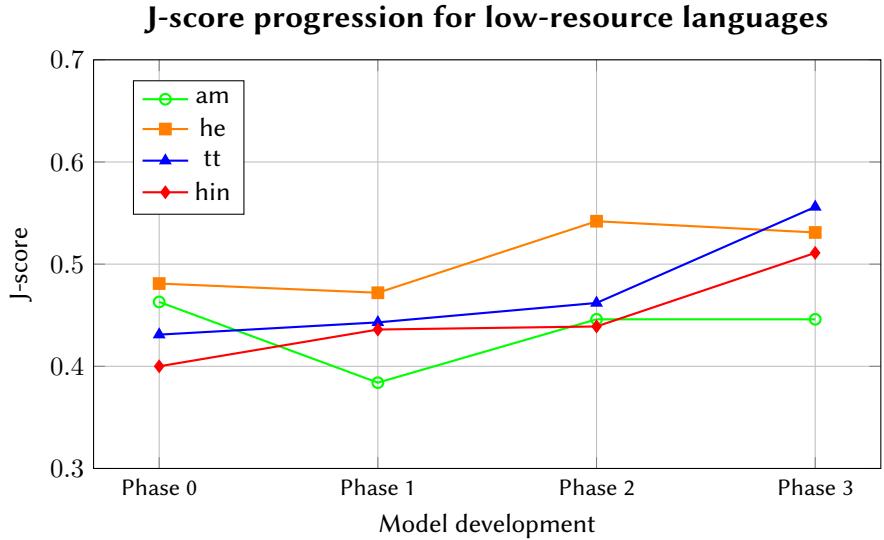
\begin{figure}[ht]
\centering
\begin{tikzpicture}
    \begin{axis}[
        width=0.75\textwidth,
        height=7cm,
        title={\large\textbf{J‐score progression for low-resource languages}},
        xlabel={Model development},
        ylabel={J‐score},
        xtick={0,1,2,3},
        xticklabels={
            {Phase 0},
            {Phase 1},
            {Phase 2},
            {Phase 3}
        },
        xticklabel style={font=\footnotesize, align=center},
        ymin=0.30, ymax=0.70,
        ytick={0.3,0.40,0.50,0.60,0.70},
        scaled y ticks=false,
        yticklabel style={/pgf/number format/fixed,
                          /pgf/number format/precision=3},
        grid=both,
        legend style={at={(0.05,0.95)}, anchor=north west, font=\small}
    ]

    \addplot[color=green,   mark=o, thick] coordinates {
        (0,0.463) (1,0.384) (2,0.446) (3,0.446)
    };
    \addlegendentry{am}

    \addplot[color=orange, mark=square*, thick] coordinates {
        (0,0.481) (1,0.472) (2,0.542) (3,0.531)
    };
    \addlegendentry{he}

    \addplot[color=blue,  mark=triangle*, thick] coordinates {
        (0,0.431) (1,0.443) (2,0.462) (3,0.556)
    };
    \addlegendentry{tt}

    \addplot[color=red,    mark=diamond*, thick] coordinates {
        (0,0.400) (1,0.436) (2,0.439) (3,0.511)
    };
    \addlegendentry{hin}
    \end{axis}
\end{tikzpicture}
\caption{J‐scores across four phases of our development of Gemma-3 12B for detoxification, specifically of low-resource and worst-achieving languages: Amharic (am), Hebrew (he), Tatar (tt), Hinglish (hin).}
\label{fig:lowres_jscore}
\end{figure}

Figure \ref{fig:lowres_jscore} reports the evolution of our joint score $J$ on the four lowest-resource languages (Amharic, Hebrew, Tatar and Hinglish) through each of the three fine-tuning phases (plus the zero-shot baseline, Phase 0). Amharic dips under few-shot CoT before modest recovery (Phase 0→3: 0.47→0.45), Hebrew gains steadily (0.48→0.53), and Tatar and Hinglish exhibit the largest rises, particularly after adding CoT and synthetic data (Tatar 0.43→0.56; Hinglish 0.40→0.51).

\subsection{Additional results from LLM-as-a-Judge}
After the end of the competition, the shared-task committee re-evaluated every submission with an LLM-as-a-Judge protocol intended to offer advanced means of analysis to the proposed models. The committee fine-tuned Llama-3.1-8B-Instruct \cite{meta2024llama31_8b_instruct} on the pairwise human annotations released for the same challenge in 2024 \cite{dementieva2024}\footnote{\url{https://github.com/textdetox/textdetox_clef_2024/tree/main/human_evaluation_results}}. The fine-tuned model outputs a non-toxicity preference score that replaces the XLM-R probability $p$ in equation (1) above (\ref{original-evaluation}), and a content-similarity score that replaces the LaBSE mixture in equation (2) above. Fluency continues to employ xCOMET as before. The joint score is therefore computed as:

\begin{equation}
\resizebox{\textwidth}{!}{$
J \;=\; X_{\mathrm{comet\_fluency}}\!\bigl(\mathrm{input}, \mathrm{output}_{\mathrm{gold}}, \mathrm{output}_{\mathrm{gen}}\bigr)
      \;\times\;
      \mathrm{SIM}_{\mathrm{LLM}}\!\bigl(\mathrm{input}, \mathrm{output}_{\mathrm{gold}}, \mathrm{output}_{\mathrm{gen}}\bigr)
      \;\times\;
      \mathrm{STA}_{\mathrm{LLM}}
$}
\end{equation}

We provide results in \ref{tab:llm-as-judge-YESPARALLEL} and \ref{tab:llm-as-judge-NOPARALLEL}: there, we include the committee's Golden Annotation performance, as well as the top two teams which ranked first in the subgroups of languages with and without the available parallel data. We additionally include the best performing baseline in each subgroup.
Unlike the original automatic evaluation where our model ranked first on several languages, the LLM-as-a-Judge protocol demotes our system to third in both subgroups. The average $J$ score across all fifteen languages would place us in second with $J=0.7669$: namely, significantly below the Golden Annotation score ($J=0.8233$) but slightly in front of the "MetaDetox" team ($J=0.7636$) and "ReText.Ai" team ($J=0.7538$).

\begin{table}[ht]
  \centering
  \caption{Results with the LLM-as-a-Judge protocol on languages with available parallel data from human annotators.}
  \label{tab:llm-as-judge-YESPARALLEL}
  \resizebox{\linewidth}{!}{%
    \begin{tabular}{lcccccccccc}
      \toprule
      Model                     & Avg J (rank) & en & es & de & zh
                                & ar & hi & uk & ru & am \\
      \midrule
      Golden annotation            & 0.820 (1)  & 0.846 & 0.783 & 0.930 & 0.716
                                & 0.838  & 0.888 & 0.807 & 0.828 & 0.742 \\
      "MetaDetox" team             & 0.812 (2)  & 0.893 & 0.823 & 0.919 & 0.813
                                & 0.826  & 0.785 & 0.791 & 0.829 & 0.626 \\
      Our model          & 0.798 (3) & 0.871 & 0.797 & 0.919 & 0.796
                                & 0.814 & 0.762 & 0.785 & 0.827 & 0.614 \\
      "ReText.Ai" team            & 0.775 (4) & 0.794 & 0.765 & 0.888 & 0.783
                                & 0.790 & 0.773 & 0.791 & 0.792 & 0.597 \\
      baseline\_mt0 & 0.768 (8) & 0.843 & 0.764 & 0.825 & 0.717
                                & 0.791 & 0.751 & 0.770 & 0.809 & 0.639 \\
      \bottomrule
    \end{tabular}%
  }
\end{table}

\begin{table}[ht]
  \caption{Results with the LLM-as-a-Judge protocol on languages with no available parallel data from human annotators.}
  \label{tab:llm-as-judge-NOPARALLEL}
  \begin{tabular}{lccccccc}
    \toprule
    Model                      & Avg J (rank) & it & ja & he
                               & fr  & tt & hin \\
    \midrule
    Golden annotation            & 0.828 (1)  & 0.893 & 0.904 & 0.783
                               & 0.724  & 0.780 & 0.887 \\
    "ReText.Ai" team             & 0.722 (2)  & 0.823 & 0.805 & 0.657
                               & 0.860  & 0.583 & 0.606 \\
    Our model          & 0.720 (3) & 0.842 & 0.820 & 0.681
                               & 0.889 & 0.495 & 0.592 \\
    "MetaDetox" team            & 0.691 (5) & 0.821 & 0.721 & 0.610
                               & 0.883 & 0.493 & 0.621 \\
    baseline\_gpt4 & 0.662 (11) & 0.790 & 0.779 & 0.578
                               & 0.865 & 0.438 & 0.524 \\
    \bottomrule
  \end{tabular}
\end{table}

The LLM acting "as judge" was trained on pairwise human preference data, and thus acts as a proxy for real-world acceptability
Our model's performance drop seems to suggest that human-calibrated LLM judgments emphasize different dimensions of text style, and a future investigation of such differences will certainly yield useful insights.
Overall, we believe that our model proved to be sufficiently robust as it maintained a competitive standing even under this additional protocol. At the same time, these results seem to reinforce the need for detoxification systems to be optimized beyond standard fluency or similarity metrics.

\subsection{Qualitative sampling analysis}
For each of the fifteen target languages, we randomly sampled and manually analyzed 150 input–output sentence pairs from the system’s submissions. Our qualitative review focused on identifying common error patterns, understanding their linguistic or resource-related drivers, and illustrating representative examples. We organize our observations into three main trends.

(i) Compound toxicity detection failures. In low- and medium-resource languages, the model frequently fails to identify toxicity when multiple toxic elements co-occur in the input. We find that this phenomenon is most pronounced in the Hinglish subset (see Appendix~\ref{appendix:D}) occurring in 6.7\% of output sentences, likely due to mixed-script and code-switching which are fairly frequent in the language corpus. We also observe similar failures in Spanish and Mandarin Chinese: regional or sociolinguistically connotated profanities are often missed, leading to incomplete detoxification\footnote{Of this type of errors, our qualitative analysis only detected one example for Spanish and one for Mandarin Chinese.}. We attribute these errors to the extensive dialectal diversity in these two languages, which, with approximately one billion and half a billion native speakers respectively, encompass a broad vocabulary of localized curse terms. Representative examples are provided in Table \ref{tab:trend1SPANISH}, where the model repeatedly failed to detect the expression "vale pico", a toxic word for “worthless” that is reported to be in use in Southern American language communities, especially in Argentina.

\begin{table}[t]
  \centering
  \caption{Examples of missed toxic elements in Spanish.}
  \label{tab:trend1SPANISH}
  \resizebox{\linewidth}{!}{%
    \begin{tabular}{p{0.10\linewidth}p{0.45\linewidth}p{0.45\linewidth}}
        \toprule
        \textbf{Language} & \textbf{Original} & \textbf{Detoxified} \\
        \midrule
        Spanish & Udec culia se demoran ene en sus trámites culiaos, vale pico! & Udec se demoran mucho en sus trámites, vale pico! \\
        Spanish & Mi perra vale pico, juro que me había levantado a jugar y tomó su peluche jajaajja Küyén de mrda & Mi novia vale pico, juro que me había levantado a jugar y tomó su peluche jajaajja Que asco. \\
        \bottomrule
    \end{tabular}
  }                            
\end{table}

(ii) Nonsensical paraphrase generation in low-resource languages. Among the lowest-resource languages, Amharic exhibits a disproportionately high rate of nonsensical outputs despite the availability of high-quality human-parallel data in the test corpus. Roughly 30\% of sampled Amharic pairs yielded ungrammatical or semantically empty sentences.

(iii) Ambiguity in toxicity definition exacerbates errors. The absence of clear-cut boundaries for “toxicity” amplifies modeling challenges, particularly for languages where the input data is dominated by political discourse. Hebrew is the clearest example: coarse annotations in the test set combine explicit insults, political disapproval, and partisan hate speech. In this context, identifying only explicit toxic terms is difficult, and models often either over-detoxify (removing non-toxic political criticism) or under-detoxify (leaving implicit hostility intact). Moreover, the fine boundaries of toxicity lead the model to substitute words that straddle polite and impolite registers (e.g., terms like “idiot”, which are common across Indo-European languages). Such substitutions result in outputs that technically remove overt insults but replace them with milder yet still demeaning vocabulary; see Table \ref{tab:trend3ambiguity}.

\begin{table}[t]
  \centering
  \caption{Examples of missed toxic elements in Spanish.}
  \label{tab:trend3ambiguity}
  \resizebox{\linewidth}{!}{%
    \begin{tabular}{p{0.10\linewidth}p{0.45\linewidth}p{0.45\linewidth}}
        \toprule
        \textbf{Language} & \textbf{Original} & \textbf{Detoxified} \\
        \midrule
        French & et en défendant comme des salopes, j'espère bien que vous gagnez ! & et en défendant comme des idiots, j'espère bien que vous gagnez ! \\
        \addlinespace
        Italian & Ora ve lo dico veramente: VAFFANCULO A TUTTI VOI. Siete solo degli sfigati: bannatemi pure ma non me ne fotte un cazzo, coglioni!!!! & Ora ve lo dico veramente: Sono arrabbiato con tutti voi. Siete solo dei perdenti: bannatemi pure ma non me ne importa niente, idioti!!!! \\
        \bottomrule
    \end{tabular}
  }                            
\end{table}

\subsection{ANOVA analysis}
\label{anova_paragraph}
We performed one-way ANOVA on mean detoxification scores across 15 languages using four grouping schemes: genetic affiliation, typology, geographical proximity, and language-resource status. Our methodology and criteria are outlined in Appendix~\ref{appendix:B}. Singleton-language clusters were aggregated into a unified “Other” category or omitted to ensure each factor level contained at least two members. Table~\ref{tab:anova-test} summarizes the Test-set results.

\begin{table*}
\caption{ANOVA: summary and effect sizes on Test set}
\label{tab:anova-test}
\begin{tabular}{lcccc}
\toprule
Method & $\eta^2$ & F-statistic & df (between, within) & $p$-value \\
\midrule
Genetic affiliation & 0.592 & 8.717 & (2, 12) & 0.00459 \\
Typology & 0.654 & 3.403 & (5, 9) & 0.05303 \\
Geographical proximity & 0.441 & 1.772 & (4, 9) & 0.21845 \\
Language-resource tier & 0.667 & 12.040 & (2, 12) & 0.00135 \\
\bottomrule
\end{tabular}
\end{table*}

The ANOVA shows that language-resource status is the strongest predictor of detox performance, explaining two-thirds of the variance ($\eta^2 = 0.667$, $F(2,12)=12.04$, $p=0.00135$). Genetic affiliation also has a statistically significant effect, accounting for nearly 60\% of variance ($\eta^2 = 0.592$, $F(2,12)=8.72$, $p=0.00459$), indicating that shared language family membership correlates with similar detox behavior. Morphological typology exhibits a large effect size ($\eta^2 = 0.654$) but narrowly fails to reach the 5\% significance threshold ($F(5,9)=3.40$, $p=0.053$), suggesting a strong trend that may warrant further study\footnote{Modern historical linguistics has largely shown how genetic affiliation does not yield typological alignment since, diachronically, languages often fail to retain their core grammatical and structural features.}. By contrast, geographical proximity shows a weaker, non-significant effect ($\eta^2 = 0.441$, $F(4,9)=1.77$, $p=0.218$), implying that regional grouping alone does not reliably predict detox performance once other factors are accounted for.

\section{Limitations}
Although fine‑tuning and curated data availability are essential, our system still largely depends on the pretrained model’s base capabilities; in the automatic evaluation phase, the zero‑shot mT0 baseline scores 0.675 on the joint metric compared to our system’s 0.685, showing only modest improvements from task‑specific adaptation.

While prompts were manually translated and validated, ensuring perfect cross-linguistic consistency remains challenging. As pointed by one of our anonymous reviewers, subtle semantic shifts in phrasing may skew model behavior across languages. Additionally, the LaBSE-based retrieval of semantically similar sentences (\ref{sec:cleaning}) can be unreliable for very low-resource languages and can likely compromise the quality of the few-shot examples.

An anonymous reviewer also rightly noted that our vast reliance on synthetic data (\ref{datasets-section}) may introduce semantic bias and encourage stylistic overfitting. Synthetic text clearly lack the diversity of human-authored examples and likely limits the robustness and generalizability of the fine-tuned system.

Furthermore, on bias specifically, we acknowledge that like most NLP systems of the same type, our detoxification model does not perform equally across social-media posts from diverse cultural, religious, or socioeconomic domains. Performance disparities arise from several sources of bias. First, our system builds on an opaque foundation model (Gemma3), whose pre-training corpus is neither public nor documented. Any skew toward well-resourced language varieties is locked into the representations we start from, and there is no fully reliable way to audit, retrace, or rebalance those original samples. Unlike tasks where data can be recollected, this selection bias is structural and irreversible: we can only mitigate downstream effects but cannot avoid bias at its source.

In fact, our qualitative review already identified missed dialectal expressions; for example, Southern-American Spanish slang “vale pico” was systematically undetected. Overall, we believe that robust, consistent text detoxification across cultural, religious, socioeconomic, and other demographics requires bias-aware adaptations at both the model pre-training, post-training, and evaluation stages.

\section{Future directions}
The well-defined evaluation metrics of detoxification tasks make them ideal candidates for preference-based optimization techniques such as GRPO and DPO, which directly optimize non-toxicity while preserving meaning. Future research could explore integrating such approaches with multilingual LLMs to fine-tune systems more effectively in line with human preferences.
Furthermore, to address the growing challenge of creative obfuscation (e.g., masked profanities like “f@ck,” “@ss” as well as emoji-based strategies) and emergent slang, especially in low-resource settings, detoxification systems should incorporate dynamic lexicon updating and character-substitution detection.

We also see promise in a multi-stage reasoning pipeline of the following type: binary toxicity classification → toxicity-type prediction → targeted detoxification → quality verification. It will be particularly interesting to explore such a pipeline and the emerging reasoning language models (RLMs) as they become more accessible to the research community.
Additionally, domain-informed data augmentation, which generates synthetic training examples that closely mirror test-time phenomena, might offer a practical path to more robust and generalizable fine-tuning corpora.

Lastly, our ANOVA results suggest that both morphological typology ($\eta^2$ = 0.654) and genetic affiliation ($\eta^2$ = 0.592) together explain the majority of performance variance in detoxification tasks. However, broader language sampling and mixed-effects modeling are necessary to disentangle their actual contributions more clearly.

\begin{acknowledgments}
  The authors would like to thank the Faculty members and all fellow students of the Master of Science program in "IT \& Cognition" at the University of Copenhagen, for their feedback and useful suggestions. We are especially grateful to Professor Patrizia Paggio and Professor Manex Agirrezabal for introducing us to the field of automatic text detoxification.
\end{acknowledgments}

\section*{Declaration on Generative AI}
 During the preparation of this work, the authors used Generative AI assistance tools for: grammar and spelling review; synthetic data generation (as outlined above in the Methods section). The authors reviewed and edited the content as needed and take full responsibility for the publication’s content.

\bibliography{sample-ceur.bib}

\begin{thebibliography}{34}
\expandafter\ifx\csname natexlab\endcsname\relax\def\natexlab#1{#1}\fi
\providecommand{\url}[1]{\texttt{#1}}
\providecommand{\href}[2]{#2}
\providecommand{\path}[1]{#1}
\providecommand{\DOIprefix}{doi:}
\providecommand{\ArXivprefix}{arXiv:}
\providecommand{\URLprefix}{URL: }
\providecommand{\Pubmedprefix}{pmid:}
\providecommand{\doi}[1]{\href{http://dx.doi.org/#1}{\path{#1}}}
\providecommand{\Pubmed}[1]{\href{pmid:#1}{\path{#1}}}
\providecommand{\bibinfo}[2]{#2}
\ifx\xfnm\relax \def\xfnm[#1]{\unskip,\space#1}\fi
\bibitem[{Bevendorff et~al.(2025)Bevendorff, Dementieva, Fr{\"o}be, Gipp, Greiner-Petter, Karlgren, Mayerl, Nakov, Panchenko, Potthast, Shelmanov, Stamatatos, Stein, Wang, Wiegmann, and Zangerle}]{bevendorff:2025}
\bibinfo{author}{J.~Bevendorff}, \bibinfo{author}{D.~Dementieva}, \bibinfo{author}{M.~Fr{\"o}be}, \bibinfo{author}{B.~Gipp}, \bibinfo{author}{A.~Greiner-Petter}, \bibinfo{author}{J.~Karlgren}, \bibinfo{author}{M.~Mayerl}, \bibinfo{author}{P.~Nakov}, \bibinfo{author}{A.~Panchenko}, \bibinfo{author}{M.~Potthast}, \bibinfo{author}{A.~Shelmanov}, \bibinfo{author}{E.~Stamatatos}, \bibinfo{author}{B.~Stein}, \bibinfo{author}{Y.~Wang}, \bibinfo{author}{M.~Wiegmann}, \bibinfo{author}{E.~Zangerle},
\newblock \bibinfo{title}{{Overview of PAN 2025: Voight-Kampff Generative AI Detection, Multilingual Text Detoxification, Multi-Author Writing Style Analysis, and Generative Plagiarism Detection}},
\newblock in: \bibinfo{editor}{J.~C. de~Albornoz}, \bibinfo{editor}{J.~Gonzalo}, \bibinfo{editor}{L.~Plaza}, \bibinfo{editor}{A.~G.~S. de~Herrera}, \bibinfo{editor}{J.~Mothe}, \bibinfo{editor}{F.~Piroi}, \bibinfo{editor}{P.~Rosso}, \bibinfo{editor}{D.~Spina}, \bibinfo{editor}{G.~Faggioli}, \bibinfo{editor}{N.~Ferro} (Eds.), \bibinfo{booktitle}{Experimental IR Meets Multilinguality, Multimodality, and Interaction. Proceedings of the Sixteenth International Conference of the CLEF Association (CLEF 2025)}, Lecture Notes in Computer Science, \bibinfo{publisher}{Springer}, \bibinfo{address}{Berlin Heidelberg New York}, \bibinfo{year}{2025}.
\bibitem[{Dementieva et~al.(2025)Dementieva, Protasov, Babakov, Rizwan, Alimova, Brune, Konovalov, Muti, Liebeskind, Litvak, Nozza, Shah~Khan, Takeshita, Vanetik, Ayele, Schneider, Wang, Yimam, Elnagar, Mukherjee, and Panchenko}]{dementieva2025overview}
\bibinfo{author}{D.~Dementieva}, \bibinfo{author}{V.~Protasov}, \bibinfo{author}{N.~Babakov}, \bibinfo{author}{N.~Rizwan}, \bibinfo{author}{I.~Alimova}, \bibinfo{author}{C.~Brune}, \bibinfo{author}{V.~Konovalov}, \bibinfo{author}{A.~Muti}, \bibinfo{author}{C.~Liebeskind}, \bibinfo{author}{M.~Litvak}, \bibinfo{author}{D.~Nozza}, \bibinfo{author}{S.~Shah~Khan}, \bibinfo{author}{S.~Takeshita}, \bibinfo{author}{N.~Vanetik}, \bibinfo{author}{A.~A. Ayele}, \bibinfo{author}{F.~Schneider}, \bibinfo{author}{X.~Wang}, \bibinfo{author}{S.~M. Yimam}, \bibinfo{author}{A.~Elnagar}, \bibinfo{author}{A.~Mukherjee}, \bibinfo{author}{A.~Panchenko},
\newblock \bibinfo{title}{Overview of the multilingual text detoxification task at pan 2025},
\newblock in: \bibinfo{editor}{G.~Faggioli}, \bibinfo{editor}{N.~Ferro}, \bibinfo{editor}{P.~Rosso}, \bibinfo{editor}{D.~Spina} (Eds.), \bibinfo{booktitle}{Working Notes of CLEF 2025 -- Conference and Labs of the Evaluation Forum}, CEUR Workshop Proceedings, \bibinfo{publisher}{CEUR-WS.org}, \bibinfo{year}{2025}.
\bibitem[{Pascale(2019)}]{Pascale19}
\bibinfo{author}{C.-M. Pascale},
\newblock \bibinfo{title}{The weaponization of language: Discourses of rising right-wing authoritarianism},
\newblock \bibinfo{journal}{Current Sociology} \bibinfo{volume}{67} (\bibinfo{year}{2019}). \DOIprefix\doi{10.1177/0011392119869963}, \bibinfo{note}{(Original work published 2019)}.
\bibitem[{Dementieva et~al.(2024)Dementieva, Moskovskiy, Babakov, Ayele, Rizwan, Schneider, Wang, Yimam, Ustalov, Stakovskii, Smirnova, Elnagar, Mukherjee, and Panchenko}]{dementieva2024}
\bibinfo{author}{D.~Dementieva}, \bibinfo{author}{D.~Moskovskiy}, \bibinfo{author}{N.~Babakov}, \bibinfo{author}{A.~A. Ayele}, \bibinfo{author}{N.~Rizwan}, \bibinfo{author}{F.~Schneider}, \bibinfo{author}{X.~Wang}, \bibinfo{author}{S.~M. Yimam}, \bibinfo{author}{D.~Ustalov}, \bibinfo{author}{E.~Stakovskii}, \bibinfo{author}{A.~Smirnova}, \bibinfo{author}{A.~Elnagar}, \bibinfo{author}{A.~Mukherjee}, \bibinfo{author}{A.~Panchenko},
\newblock \bibinfo{title}{Overview of the multilingual text detoxification task at pan 2024},
\newblock in: \bibinfo{booktitle}{Working Notes of the Conference and Labs of the Evaluation Forum (CLEF 2024)}, volume \bibinfo{volume}{3740} of \textit{\bibinfo{series}{CEUR Workshop Proceedings}}, \bibinfo{publisher}{CEUR-WS.org}, \bibinfo{address}{Grenoble, France}, \bibinfo{year}{2024}, pp. \bibinfo{pages}{2432--2461}.
\bibitem[{jig(2018)}]{jigsaw2018toxiccomment}
\bibinfo{title}{Toxic comment classification challenge dataset}, \bibinfo{howpublished}{Kaggle competition dataset}, \bibinfo{year}{2018}. \URLprefix \url{https://www.kaggle.com/competitions/jigsaw-toxic-comment-classification-challenge}.
\bibitem[{Nogueira~dos Santos et~al.(2018)Nogueira~dos Santos, Melnyk, and Padhi}]{nogueira-dos-santos-etal-2018-fighting}
\bibinfo{author}{C.~Nogueira~dos Santos}, \bibinfo{author}{I.~Melnyk}, \bibinfo{author}{I.~Padhi},
\newblock \bibinfo{title}{Fighting offensive language on social media with unsupervised text style transfer},
\newblock in: \bibinfo{editor}{I.~Gurevych}, \bibinfo{editor}{Y.~Miyao} (Eds.), \bibinfo{booktitle}{Proceedings of the 56th Annual Meeting of the Association for Computational Linguistics (Volume 2: Short Papers)}, \bibinfo{publisher}{Association for Computational Linguistics}, \bibinfo{address}{Melbourne, Australia}, \bibinfo{year}{2018}, pp. \bibinfo{pages}{189--194}. \URLprefix \url{https://aclanthology.org/P18-2031/}. \DOIprefix\doi{10.18653/v1/P18-2031}.
\bibitem[{Dale et~al.(2021)Dale, Voronov, Dementieva, Logacheva, Kozlova, Semenov, and Panchenko}]{dale-etal-2021-text}
\bibinfo{author}{D.~Dale}, \bibinfo{author}{A.~Voronov}, \bibinfo{author}{D.~Dementieva}, \bibinfo{author}{V.~Logacheva}, \bibinfo{author}{O.~Kozlova}, \bibinfo{author}{N.~Semenov}, \bibinfo{author}{A.~Panchenko},
\newblock \bibinfo{title}{Text detoxification using large pre-trained neural models},
\newblock in: \bibinfo{editor}{M.-F. Moens}, \bibinfo{editor}{X.~Huang}, \bibinfo{editor}{L.~Specia}, \bibinfo{editor}{S.~W.-t. Yih} (Eds.), \bibinfo{booktitle}{Proceedings of the 2021 Conference on Empirical Methods in Natural Language Processing}, \bibinfo{publisher}{Association for Computational Linguistics}, \bibinfo{address}{Online and Punta Cana, Dominican Republic}, \bibinfo{year}{2021}, pp. \bibinfo{pages}{7979--7996}. \URLprefix \url{https://aclanthology.org/2021.emnlp-main.629/}. \DOIprefix\doi{10.18653/v1/2021.emnlp-main.629}.
\bibitem[{Logacheva et~al.(2022)Logacheva, Dementieva, Ustyantsev, Moskovskiy, Dale, Krotova, Semenov, and Panchenko}]{logacheva-etal-2022-paradetox}
\bibinfo{author}{V.~Logacheva}, \bibinfo{author}{D.~Dementieva}, \bibinfo{author}{S.~Ustyantsev}, \bibinfo{author}{D.~Moskovskiy}, \bibinfo{author}{D.~Dale}, \bibinfo{author}{I.~Krotova}, \bibinfo{author}{N.~Semenov}, \bibinfo{author}{A.~Panchenko},
\newblock \bibinfo{title}{Paradetox: Detoxification with parallel data},
\newblock in: \bibinfo{editor}{S.~Muresan}, \bibinfo{editor}{P.~Nakov}, \bibinfo{editor}{A.~Villavicencio} (Eds.), \bibinfo{booktitle}{Proceedings of the 60th Annual Meeting of the Association for Computational Linguistics (Volume 1: Long Papers)}, \bibinfo{publisher}{Association for Computational Linguistics}, \bibinfo{address}{Dublin, Ireland}, \bibinfo{year}{2022}, pp. \bibinfo{pages}{6804--6818}. \URLprefix \url{https://aclanthology.org/2022.acl-long.469/}. \DOIprefix\doi{10.18653/v1/2022.acl-long.469}.
\bibitem[{Pesaranghader et~al.(2024)Pesaranghader, Verma, and Bharadwaj}]{pesaranghader2024gptdetoxincontextlearningbasedparaphraser}
\bibinfo{author}{A.~Pesaranghader}, \bibinfo{author}{N.~Verma}, \bibinfo{author}{M.~Bharadwaj}, \bibinfo{title}{Gpt-detox: An in-context learning-based paraphraser for text detoxification}, \bibinfo{year}{2024}. \URLprefix \url{https://arxiv.org/abs/2404.03052}. \DOIprefix\doi{10.1109/ICMLA58977.2023.00230}. \href{http://arxiv.org/abs/2404.03052}{{\tt arXiv:2404.03052}}.
\bibitem[{Dementieva et~al.(2024)Dementieva, Babakov, and Panchenko}]{dementieva-etal-2024-multiparadetox}
\bibinfo{author}{D.~Dementieva}, \bibinfo{author}{N.~Babakov}, \bibinfo{author}{A.~Panchenko},
\newblock \bibinfo{title}{Multiparadetox: Extending text detoxification with parallel data to new languages},
\newblock in: \bibinfo{editor}{K.~Duh}, \bibinfo{editor}{H.~Gomez}, \bibinfo{editor}{S.~Bethard} (Eds.), \bibinfo{booktitle}{Proceedings of the 2024 Conference of the North American Chapter of the Association for Computational Linguistics: Human Language Technologies (Volume 2: Short Papers)}, \bibinfo{publisher}{Association for Computational Linguistics}, \bibinfo{address}{Mexico City, Mexico}, \bibinfo{year}{2024}, pp. \bibinfo{pages}{124--140}. \URLprefix \url{https://aclanthology.org/2024.naacl-short.12/}. \DOIprefix\doi{10.18653/v1/2024.naacl-short.12}.
\bibitem[{Teo(2023)}]{teo2023costar}
\bibinfo{author}{S.~Teo}, \bibinfo{title}{The co-star framework for prompt engineering}, \bibinfo{howpublished}{Blog post, *Towards Data Science*}, \bibinfo{year}{2023}. \URLprefix \url{https://towardsdatascience.com/how-i-won-singapores-gpt-4-prompt-engineering-competition-34c195a93d41}.
\bibitem[{Dementieva et~al.(2025)Dementieva, Babakov, Ronen, Ayele, Rizwan, Schneider, Wang, Yimam, Moskovskiy, Stakovskii, Kaufman, Elnagar, Mukherjee, and Panchenko}]{dementieva-etal-2025-multilingual}
\bibinfo{author}{D.~Dementieva}, \bibinfo{author}{N.~Babakov}, \bibinfo{author}{A.~Ronen}, \bibinfo{author}{A.~A. Ayele}, \bibinfo{author}{N.~Rizwan}, \bibinfo{author}{F.~Schneider}, \bibinfo{author}{X.~Wang}, \bibinfo{author}{S.~M. Yimam}, \bibinfo{author}{D.~Moskovskiy}, \bibinfo{author}{E.~Stakovskii}, \bibinfo{author}{E.~Kaufman}, \bibinfo{author}{A.~Elnagar}, \bibinfo{author}{A.~Mukherjee}, \bibinfo{author}{A.~Panchenko},
\newblock \bibinfo{title}{Multilingual and explainable text detoxification with parallel corpora},
\newblock in: \bibinfo{editor}{O.~Rambow}, \bibinfo{editor}{L.~Wanner}, \bibinfo{editor}{M.~Apidianaki}, \bibinfo{editor}{H.~Al-Khalifa}, \bibinfo{editor}{B.~Di~Eugenio}, \bibinfo{editor}{S.~Schockaert} (Eds.), \bibinfo{booktitle}{Proceedings of the 31st International Conference on Computational Linguistics}, \bibinfo{publisher}{Association for Computational Linguistics}, \bibinfo{address}{Abu Dhabi, UAE}, \bibinfo{year}{2025}, pp. \bibinfo{pages}{7998--8025}. \URLprefix \url{https://aclanthology.org/2025.coling-main.535/}.
\bibitem[{Floto et~al.(2023)Floto, Abdollah~Pour, Farinneya, Tang, Pesaranghader, Bharadwaj, and Sanner}]{floto2023diffudetoxmixeddiffusionmodel}
\bibinfo{author}{G.~Floto}, \bibinfo{author}{M.~M. Abdollah~Pour}, \bibinfo{author}{P.~Farinneya}, \bibinfo{author}{Z.~Tang}, \bibinfo{author}{A.~Pesaranghader}, \bibinfo{author}{M.~Bharadwaj}, \bibinfo{author}{S.~Sanner}, \bibinfo{title}{Diffudetox: A mixed diffusion model for text detoxification}, \bibinfo{year}{2023}. \URLprefix \url{https://arxiv.org/abs/2306.08505}. \href{http://arxiv.org/abs/2306.08505}{{\tt arXiv:2306.08505}}.
\bibitem[{Shah(2022)}]{shah2022_rlm_hinglish_translator}
\bibinfo{author}{R.~Shah}, \bibinfo{title}{Rlm--hinglish translator}, \bibinfo{howpublished}{\url{https://huggingface.co/rudrashah/RLM-hinglish-translator}}, \bibinfo{year}{2022}. \bibinfo{note}{Accessed: 2025-05-28}.
\bibitem[{{NLLB Team}(2022)}]{nllbteam2022nllb}
\bibinfo{author}{{NLLB Team}},
\newblock \bibinfo{title}{{No Language Left Behind}: Building a scalable and inclusive multilingual translation system},
\newblock \bibinfo{journal}{arXiv preprint arXiv:2207.04672}  (\bibinfo{year}{2022}).
\bibitem[{Dang and contributors(2025)}]{anhdtd2025_paradetox_mt6}
\bibinfo{author}{A.~D. Dang}, \bibinfo{author}{contributors}, \bibinfo{title}{Paradetox–mt6: Synthetic multilingual detoxification corpus}, \bibinfo{howpublished}{\url{https://huggingface.co/datasets/anhdtd/augument_data}}, \bibinfo{year}{2025}. \bibinfo{note}{Version 1.0, accessed 28 May 2025}.
\bibitem[{{TextDetox Team}(2024)}]{textdetox2024_multilingual_toxicity}
\bibinfo{author}{{TextDetox Team}}, \bibinfo{title}{Multilingual toxicity dataset}, \bibinfo{howpublished}{\url{https://huggingface.co/datasets/textdetox/multilingual_toxicity_dataset}}, \bibinfo{year}{2024}. \bibinfo{note}{Version 1.0, accessed 28~May~2025}.
\bibitem[{Moskovskiy et~al.(2025)Moskovskiy, Sushko, Pletenev, Tutubalina, and Panchenko}]{moskovskiy2025synthdetoxmmodernllmsfewshot}
\bibinfo{author}{D.~Moskovskiy}, \bibinfo{author}{N.~Sushko}, \bibinfo{author}{S.~Pletenev}, \bibinfo{author}{E.~Tutubalina}, \bibinfo{author}{A.~Panchenko}, \bibinfo{title}{Synthdetoxm: Modern llms are few-shot parallel detoxification data annotators}, \bibinfo{year}{2025}. \URLprefix \url{https://arxiv.org/abs/2502.06394}. \href{http://arxiv.org/abs/2502.06394}{{\tt arXiv:2502.06394}}.
\bibitem[{Zeira and Montal(2024)}]{zeira2024_multilingual_toxic_lexicon}
\bibinfo{author}{A.~Zeira}, \bibinfo{author}{J.~Montal}, \bibinfo{title}{Multilingual toxic lexicon}, \bibinfo{howpublished}{Hugging Face dataset}, \bibinfo{year}{2024}. \URLprefix \url{https://huggingface.co/datasets/textdetox/multilingual_toxic_lexicon/}, \bibinfo{note}{accessed: 2025-05-28}.
\bibitem[{{NLLB Team}(2023)}]{nllb2023_distilled_600m}
\bibinfo{author}{{NLLB Team}}, \bibinfo{title}{facebook/nllb-200-distilled-600m: A distilled subset of nllb-200}, \bibinfo{howpublished}{Hugging Face model}, \bibinfo{year}{2023}. \URLprefix \url{https://huggingface.co/facebook/nllb-200-distilled-600M}, \bibinfo{note}{accessed: 2025-05-28}.
\bibitem[{{S-NLP Research}(2023)}]{snlp2023_bart_base_detox}
\bibinfo{author}{{S-NLP Research}}, \bibinfo{title}{s-nlp/bart-base-detox: Bart-base model fine-tuned for english detoxification}, \bibinfo{howpublished}{Hugging Face model}, \bibinfo{year}{2023}. \URLprefix \url{https://huggingface.co/s-nlp/bart-base-detox}, \bibinfo{note}{accessed: 2025-05-28}.
\bibitem[{Labonne(2024)}]{labonne2024_llama_instruct}
\bibinfo{author}{M.~Labonne}, \bibinfo{title}{mlabonne/llama-3.1-70b-instruct-lorablated: Llama-3.1 70b instruction-tuned variant}, \bibinfo{howpublished}{Hugging Face model}, \bibinfo{year}{2024}. \URLprefix \url{https://huggingface.co/mlabonne/Llama-3.1-70B-Instruct-lorablated}, \bibinfo{note}{accessed: 2025-05-28}.
\bibitem[{OpenAI(2025)}]{openai2025_gpt_models}
\bibinfo{author}{OpenAI}, \bibinfo{title}{Gpt-4-0613, gpt-4o-2024-08-06, and o3-mini-2025-01-31: proprietary large language models}, \bibinfo{howpublished}{OpenAI API}, \bibinfo{year}{2025}. \bibinfo{note}{Accessed: 2025-05-28}.
\bibitem[{Team et~al.(2025)Team, Kamath, Ferret, Pathak, Vieillard, Merhej, Perrin, Matejovicova, Ramé, Rivière, Rouillard, Mesnard, Cideron, bastien Grill, Ramos, Yvinec, Casbon, Pot, Penchev, Liu, Visin, Kenealy, Beyer, Zhai, Tsitsulin, Busa-Fekete, Feng, Sachdeva, Coleman, Gao, Mustafa, Barr, Parisotto, Tian, Eyal, Cherry, Peter, Sinopalnikov, Bhupatiraju, Agarwal, Kazemi, Malkin, Kumar, Vilar, Brusilovsky, Luo, Steiner, Friesen, Sharma, Sharma, Gilady, Goedeckemeyer, Saade, Feng, Kolesnikov, Bendebury, Abdagic, Vadi, György, Pinto, Das, Bapna, Miech, Yang, Paterson, Shenoy, Chakrabarti, Piot, Wu, Shahriari, Petrini, Chen, Lan, Choquette-Choo, Carey, Brick, Deutsch, Eisenbud, Cattle, Cheng, Paparas, Sreepathihalli, Reid, Tran, Zelle, Noland, Huizenga, Kharitonov, Liu, Amirkhanyan, Cameron, Hashemi, Klimczak-Plucińska, Singh, Mehta, Lehri, Hazimeh, Ballantyne, Szpektor, Nardini, Pouget-Abadie, Chan, Stanton, Wieting, Lai, Orbay, Fernandez, Newlan, yeong Ji, Singh, Black, Yu, Hui, Vodrahalli, Greff, Qiu,
  Valentine, Coelho, Ritter, Hoffman, Watson, Chaturvedi, Moynihan, Ma, Babar, Noy, Byrd, Roy, Momchev, Chauhan, Sachdeva, Bunyan, Botarda, Caron, Rubenstein, Culliton, Schmid, Sessa, Xu, Stanczyk, Tafti, Shivanna, Wu, Pan, Rokni, Willoughby, Vallu, Mullins, Jerome, Smoot, Girgin, Iqbal, Reddy, Sheth, Põder, Bhatnagar, Panyam, Eiger, Zhang, Liu, Yacovone, Liechty, Kalra, Evci, Misra, Roseberry, Feinberg, Kolesnikov, Han, Kwon, Chen, Chow, Zhu, Wei, Egyed, Cotruta, Giang, Kirk, Rao, Black, Babar, Lo, Moreira, Martins, Sanseviero, Gonzalez, Gleicher, Warkentin, Mirrokni, Senter, Collins, Barral, Ghahramani, Hadsell, Matias, Sculley, Petrov, Fiedel, Shazeer, Vinyals, Dean, Hassabis, Kavukcuoglu, Farabet, Buchatskaya, Alayrac, Anil, Dmitry, Lepikhin, Borgeaud, Bachem, Joulin, Andreev, Hardin, Dadashi, and Hussenot}]{gemma3technicalreport}
\bibinfo{author}{G.~Team}, \bibinfo{author}{A.~Kamath}, \bibinfo{author}{J.~Ferret}, \bibinfo{author}{S.~Pathak}, \bibinfo{author}{N.~Vieillard}, \bibinfo{author}{R.~Merhej}, \bibinfo{author}{S.~Perrin}, \bibinfo{author}{T.~Matejovicova}, \bibinfo{author}{A.~Ramé}, \bibinfo{author}{M.~Rivière}, \bibinfo{author}{L.~Rouillard}, \bibinfo{author}{T.~Mesnard}, \bibinfo{author}{G.~Cideron}, \bibinfo{author}{J.~bastien Grill}, \bibinfo{author}{S.~Ramos}, \bibinfo{author}{E.~Yvinec}, \bibinfo{author}{M.~Casbon}, \bibinfo{author}{E.~Pot}, \bibinfo{author}{I.~Penchev}, \bibinfo{author}{G.~Liu}, \bibinfo{author}{F.~Visin}, \bibinfo{author}{K.~Kenealy}, \bibinfo{author}{L.~Beyer}, \bibinfo{author}{X.~Zhai}, \bibinfo{author}{A.~Tsitsulin}, \bibinfo{author}{R.~Busa-Fekete}, \bibinfo{author}{A.~Feng}, \bibinfo{author}{N.~Sachdeva}, \bibinfo{author}{B.~Coleman}, \bibinfo{author}{Y.~Gao}, \bibinfo{author}{B.~Mustafa}, \bibinfo{author}{I.~Barr}, \bibinfo{author}{E.~Parisotto}, \bibinfo{author}{D.~Tian},
  \bibinfo{author}{M.~Eyal}, \bibinfo{author}{C.~Cherry}, \bibinfo{author}{J.-T. Peter}, \bibinfo{author}{D.~Sinopalnikov}, \bibinfo{author}{S.~Bhupatiraju}, \bibinfo{author}{R.~Agarwal}, \bibinfo{author}{M.~Kazemi}, \bibinfo{author}{D.~Malkin}, \bibinfo{author}{R.~Kumar}, \bibinfo{author}{D.~Vilar}, \bibinfo{author}{I.~Brusilovsky}, \bibinfo{author}{J.~Luo}, \bibinfo{author}{A.~Steiner}, \bibinfo{author}{A.~Friesen}, \bibinfo{author}{A.~Sharma}, \bibinfo{author}{A.~Sharma}, \bibinfo{author}{A.~M. Gilady}, \bibinfo{author}{A.~Goedeckemeyer}, \bibinfo{author}{A.~Saade}, \bibinfo{author}{A.~Feng}, \bibinfo{author}{A.~Kolesnikov}, \bibinfo{author}{A.~Bendebury}, \bibinfo{author}{A.~Abdagic}, \bibinfo{author}{A.~Vadi}, \bibinfo{author}{A.~György}, \bibinfo{author}{A.~S. Pinto}, \bibinfo{author}{A.~Das}, \bibinfo{author}{A.~Bapna}, \bibinfo{author}{A.~Miech}, \bibinfo{author}{A.~Yang}, \bibinfo{author}{A.~Paterson}, \bibinfo{author}{A.~Shenoy}, \bibinfo{author}{A.~Chakrabarti}, \bibinfo{author}{B.~Piot},
  \bibinfo{author}{B.~Wu}, \bibinfo{author}{B.~Shahriari}, \bibinfo{author}{B.~Petrini}, \bibinfo{author}{C.~Chen}, \bibinfo{author}{C.~L. Lan}, \bibinfo{author}{C.~A. Choquette-Choo}, \bibinfo{author}{C.~Carey}, \bibinfo{author}{C.~Brick}, \bibinfo{author}{D.~Deutsch}, \bibinfo{author}{D.~Eisenbud}, \bibinfo{author}{D.~Cattle}, \bibinfo{author}{D.~Cheng}, \bibinfo{author}{D.~Paparas}, \bibinfo{author}{D.~S. Sreepathihalli}, \bibinfo{author}{D.~Reid}, \bibinfo{author}{D.~Tran}, \bibinfo{author}{D.~Zelle}, \bibinfo{author}{E.~Noland}, \bibinfo{author}{E.~Huizenga}, \bibinfo{author}{E.~Kharitonov}, \bibinfo{author}{F.~Liu}, \bibinfo{author}{G.~Amirkhanyan}, \bibinfo{author}{G.~Cameron}, \bibinfo{author}{H.~Hashemi}, \bibinfo{author}{H.~Klimczak-Plucińska}, \bibinfo{author}{H.~Singh}, \bibinfo{author}{H.~Mehta}, \bibinfo{author}{H.~T. Lehri}, \bibinfo{author}{H.~Hazimeh}, \bibinfo{author}{I.~Ballantyne}, \bibinfo{author}{I.~Szpektor}, \bibinfo{author}{I.~Nardini}, \bibinfo{author}{J.~Pouget-Abadie},
  \bibinfo{author}{J.~Chan}, \bibinfo{author}{J.~Stanton}, \bibinfo{author}{J.~Wieting}, \bibinfo{author}{J.~Lai}, \bibinfo{author}{J.~Orbay}, \bibinfo{author}{J.~Fernandez}, \bibinfo{author}{J.~Newlan}, \bibinfo{author}{J.~yeong Ji}, \bibinfo{author}{J.~Singh}, \bibinfo{author}{K.~Black}, \bibinfo{author}{K.~Yu}, \bibinfo{author}{K.~Hui}, \bibinfo{author}{K.~Vodrahalli}, \bibinfo{author}{K.~Greff}, \bibinfo{author}{L.~Qiu}, \bibinfo{author}{M.~Valentine}, \bibinfo{author}{M.~Coelho}, \bibinfo{author}{M.~Ritter}, \bibinfo{author}{M.~Hoffman}, \bibinfo{author}{M.~Watson}, \bibinfo{author}{M.~Chaturvedi}, \bibinfo{author}{M.~Moynihan}, \bibinfo{author}{M.~Ma}, \bibinfo{author}{N.~Babar}, \bibinfo{author}{N.~Noy}, \bibinfo{author}{N.~Byrd}, \bibinfo{author}{N.~Roy}, \bibinfo{author}{N.~Momchev}, \bibinfo{author}{N.~Chauhan}, \bibinfo{author}{N.~Sachdeva}, \bibinfo{author}{O.~Bunyan}, \bibinfo{author}{P.~Botarda}, \bibinfo{author}{P.~Caron}, \bibinfo{author}{P.~K. Rubenstein}, \bibinfo{author}{P.~Culliton},
  \bibinfo{author}{P.~Schmid}, \bibinfo{author}{P.~G. Sessa}, \bibinfo{author}{P.~Xu}, \bibinfo{author}{P.~Stanczyk}, \bibinfo{author}{P.~Tafti}, \bibinfo{author}{R.~Shivanna}, \bibinfo{author}{R.~Wu}, \bibinfo{author}{R.~Pan}, \bibinfo{author}{R.~Rokni}, \bibinfo{author}{R.~Willoughby}, \bibinfo{author}{R.~Vallu}, \bibinfo{author}{R.~Mullins}, \bibinfo{author}{S.~Jerome}, \bibinfo{author}{S.~Smoot}, \bibinfo{author}{S.~Girgin}, \bibinfo{author}{S.~Iqbal}, \bibinfo{author}{S.~Reddy}, \bibinfo{author}{S.~Sheth}, \bibinfo{author}{S.~Põder}, \bibinfo{author}{S.~Bhatnagar}, \bibinfo{author}{S.~R. Panyam}, \bibinfo{author}{S.~Eiger}, \bibinfo{author}{S.~Zhang}, \bibinfo{author}{T.~Liu}, \bibinfo{author}{T.~Yacovone}, \bibinfo{author}{T.~Liechty}, \bibinfo{author}{U.~Kalra}, \bibinfo{author}{U.~Evci}, \bibinfo{author}{V.~Misra}, \bibinfo{author}{V.~Roseberry}, \bibinfo{author}{V.~Feinberg}, \bibinfo{author}{V.~Kolesnikov}, \bibinfo{author}{W.~Han}, \bibinfo{author}{W.~Kwon}, \bibinfo{author}{X.~Chen},
  \bibinfo{author}{Y.~Chow}, \bibinfo{author}{Y.~Zhu}, \bibinfo{author}{Z.~Wei}, \bibinfo{author}{Z.~Egyed}, \bibinfo{author}{V.~Cotruta}, \bibinfo{author}{M.~Giang}, \bibinfo{author}{P.~Kirk}, \bibinfo{author}{A.~Rao}, \bibinfo{author}{K.~Black}, \bibinfo{author}{N.~Babar}, \bibinfo{author}{J.~Lo}, \bibinfo{author}{E.~Moreira}, \bibinfo{author}{L.~G. Martins}, \bibinfo{author}{O.~Sanseviero}, \bibinfo{author}{L.~Gonzalez}, \bibinfo{author}{Z.~Gleicher}, \bibinfo{author}{T.~Warkentin}, \bibinfo{author}{V.~Mirrokni}, \bibinfo{author}{E.~Senter}, \bibinfo{author}{E.~Collins}, \bibinfo{author}{J.~Barral}, \bibinfo{author}{Z.~Ghahramani}, \bibinfo{author}{R.~Hadsell}, \bibinfo{author}{Y.~Matias}, \bibinfo{author}{D.~Sculley}, \bibinfo{author}{S.~Petrov}, \bibinfo{author}{N.~Fiedel}, \bibinfo{author}{N.~Shazeer}, \bibinfo{author}{O.~Vinyals}, \bibinfo{author}{J.~Dean}, \bibinfo{author}{D.~Hassabis}, \bibinfo{author}{K.~Kavukcuoglu}, \bibinfo{author}{C.~Farabet}, \bibinfo{author}{E.~Buchatskaya},
  \bibinfo{author}{J.-B. Alayrac}, \bibinfo{author}{R.~Anil}, \bibinfo{author}{Dmitry}, \bibinfo{author}{Lepikhin}, \bibinfo{author}{S.~Borgeaud}, \bibinfo{author}{O.~Bachem}, \bibinfo{author}{A.~Joulin}, \bibinfo{author}{A.~Andreev}, \bibinfo{author}{C.~Hardin}, \bibinfo{author}{R.~Dadashi}, \bibinfo{author}{L.~Hussenot}, \bibinfo{title}{Gemma 3 technical report}, \bibinfo{year}{2025}. \URLprefix \url{https://arxiv.org/abs/2503.19786}. \href{http://arxiv.org/abs/2503.19786}{{\tt arXiv:2503.19786}}.
\bibitem[{Wei et~al.(2023)Wei, Wang, Schuurmans, Bosma, Ichter, Xia, Chi, Le, and Zhou}]{wei2023chainofthoughtpromptingelicitsreasoning}
\bibinfo{author}{J.~Wei}, \bibinfo{author}{X.~Wang}, \bibinfo{author}{D.~Schuurmans}, \bibinfo{author}{M.~Bosma}, \bibinfo{author}{B.~Ichter}, \bibinfo{author}{F.~Xia}, \bibinfo{author}{E.~Chi}, \bibinfo{author}{Q.~Le}, \bibinfo{author}{D.~Zhou}, \bibinfo{title}{Chain-of-thought prompting elicits reasoning in large language models}, \bibinfo{year}{2023}. \URLprefix \url{https://arxiv.org/abs/2201.11903}. \href{http://arxiv.org/abs/2201.11903}{{\tt arXiv:2201.11903}}.
\bibitem[{Face(2023{\natexlab{a}})}]{xlmr_toxicity_classifier}
\bibinfo{author}{H.~Face}, \bibinfo{title}{xlmr-large-toxicity-classifier-v2}, \bibinfo{howpublished}{\url{https://huggingface.co/textdetox/xlmr-large-toxicity-classifier-v2}}, \bibinfo{year}{2023}{\natexlab{a}}. \bibinfo{note}{Accessed: 2025-05-28}.
\bibitem[{Face(2023{\natexlab{b}})}]{labse}
\bibinfo{author}{H.~Face}, \bibinfo{title}{Labse}, \bibinfo{howpublished}{\url{https://huggingface.co/sentence-transformers/LaBSE}}, \bibinfo{year}{2023}{\natexlab{b}}. \bibinfo{note}{Accessed: 2025-05-28}.
\bibitem[{Face(2023{\natexlab{c}})}]{xcomet_lite}
\bibinfo{author}{H.~Face}, \bibinfo{title}{Xcomet-lite}, \bibinfo{howpublished}{\url{https://huggingface.co/myyycroft/XCOMET-lite}}, \bibinfo{year}{2023}{\natexlab{c}}. \bibinfo{note}{Accessed: 2025-05-28}.
\bibitem[{Hu et~al.(2022)Hu, Shen, Wallis, Allen-Zhu, Li, Wang, Wang, and Chen}]{hu2022lora}
\bibinfo{author}{E.~J. Hu}, \bibinfo{author}{Y.~Shen}, \bibinfo{author}{P.~Wallis}, \bibinfo{author}{Z.~Allen-Zhu}, \bibinfo{author}{Y.~Li}, \bibinfo{author}{S.~Wang}, \bibinfo{author}{L.~Wang}, \bibinfo{author}{W.~Chen},
\newblock \bibinfo{title}{Lora: Low-rank adaptation of large language models},
\newblock in: \bibinfo{booktitle}{Proceedings of the Tenth International Conference on Learning Representations (ICLR)}, \bibinfo{year}{2022}. \bibinfo{note}{ArXiv:2106.09685}.
\bibitem[{{Meta AI}(2024)}]{meta2024llama31_8b_instruct}
\bibinfo{author}{{Meta AI}}, \bibinfo{title}{{Llama-3.1-8B-Instruct}}, \bibinfo{howpublished}{\url{https://huggingface.co/meta-llama/Llama-3.1-8B-Instruct}}, \bibinfo{year}{2024}. \bibinfo{note}{Model version 3.1, accessed 6 Jul 2025}.
\bibitem[{Ciancaglini(2008)}]{Ciancaglini2008}
\bibinfo{author}{C.~A. Ciancaglini},
\newblock \bibinfo{title}{How to prove genetic relationships among languages: The cases of japanese and korean},
\newblock \bibinfo{journal}{Rivista Degli Studi Orientali}  (\bibinfo{year}{2008}). \URLprefix \url{https://doi.org/10.1400/143140}. \DOIprefix\doi{10.1400/143140}.
\bibitem[{Moravcsik(2013)}]{Moravcsik2013introducing}
\bibinfo{author}{E.~A. Moravcsik}, \bibinfo{title}{Introducing Language Typology}, Cambridge Introductions to Language and Linguistics, \bibinfo{edition}{illustrated} ed., \bibinfo{publisher}{Cambridge University Press}, \bibinfo{address}{Cambridge}, \bibinfo{year}{2013}. \DOIprefix\doi{10.1017/CBO9780511978876}.
\bibitem[{Son et~al.(2025)Son, Yoon, Suk, Aula-Blasco, Aslan, Kim, Islam, Prats-Cristià, Tormo-Bañuelos, and Kim}]{son2025mmevalmultilingualmetaevaluationbenchmark}
\bibinfo{author}{G.~Son}, \bibinfo{author}{D.~Yoon}, \bibinfo{author}{J.~Suk}, \bibinfo{author}{J.~Aula-Blasco}, \bibinfo{author}{M.~Aslan}, \bibinfo{author}{V.~T. Kim}, \bibinfo{author}{S.~B. Islam}, \bibinfo{author}{J.~Prats-Cristià}, \bibinfo{author}{L.~Tormo-Bañuelos}, \bibinfo{author}{S.~Kim}, \bibinfo{title}{Mm-eval: A multilingual meta-evaluation benchmark for llm-as-a-judge and reward models}, \bibinfo{year}{2025}. \URLprefix \url{https://arxiv.org/abs/2410.17578}. \href{http://arxiv.org/abs/2410.17578}{{\tt arXiv:2410.17578}}.
\bibitem[{Zhong et~al.(2024)Zhong, Yang, Liu, Zhang, Liu, Sun, Pan, Li, Zhou, Jiang, Chen, and Liu}]{zhong2024opportunitieschallengeslargelanguage}
\bibinfo{author}{T.~Zhong}, \bibinfo{author}{Z.~Yang}, \bibinfo{author}{Z.~Liu}, \bibinfo{author}{R.~Zhang}, \bibinfo{author}{Y.~Liu}, \bibinfo{author}{H.~Sun}, \bibinfo{author}{Y.~Pan}, \bibinfo{author}{Y.~Li}, \bibinfo{author}{Y.~Zhou}, \bibinfo{author}{H.~Jiang}, \bibinfo{author}{J.~Chen}, \bibinfo{author}{T.~Liu}, \bibinfo{title}{Opportunities and challenges of large language models for low-resource languages in humanities research}, \bibinfo{year}{2024}. \URLprefix \url{https://arxiv.org/abs/2412.04497}. \href{http://arxiv.org/abs/2412.04497}{{\tt arXiv:2412.04497}}.

\end{thebibliography}

\appendix

\section{Appendix: Data augmentation through machine translation}
\begin{table}[ht]
  \caption{Summary of the machine‑translation models used for data augmentation.}
  \label{tab:mt-models}
  \centering
  \begin{tabular}{p{3cm}p{4.5cm}p{4.9cm}p{2.8cm}}
    \toprule
    \textbf{Model} & \textbf{Architecture (params)} & \textbf{Training corpus} & \textbf{Tokenizer} \\
    \midrule
    RLM--Hinglish Translator & Gemma‑2B causal LM; LoRA‑fine‑tuned & 100 M Hinglish--English sentence pairs & 64 k BPE \\
    NLLB‑200 & 24‑layer encoder--decoder Transformer; 3.3 B params & 4.4 B multilingual sentences (200 langs.) & 256 k SentencePiece \\
    \bottomrule
  \end{tabular}
\end{table}

\section{Appendix: Grouping methodology for ANOVA analysis}
\label{appendix:B}
\subsection{Genetic affiliation}
Eight Indo-European languages (English, Spanish, German, Hindi, Ukrainian, Russian, Italian, French) formed the largest cluster. The three Semitic tongues (Arabic, Amharic, Hebrew) comprised a second, genetically coherent group. Chinese stood alone as Sino-Tibetan, Tatar as Turkic. Japanese was also treated as unclassified since, unlike common misconceptions in modern historical linguistics, no definitive evidence has been found linking Japanese to any other linguistic family \cite{Ciancaglini2008}. Hinglish, a contemporary English–Hindi pidgin, was non included in the Indo-European cluster. Therefore the latter four languages (Chinese, Tatar, Hinglish, Japanese) were combined into an "Other" group of unrelated members.

\subsection{Typology}
We grouped languages by their dominant word-order and morphological processes, with the latter features being prioritized  \cite{Moravcsik2013introducing}. The “fusional + SVO” cluster included the Western Romance languages (Spanish, Italian, French). Semitic root-and-pattern morphology defined the “fusional + templatic” group (Arabic, Amharic, Hebrew). The case-rich fusional-synthetic category (of "case-languages") comprised Ukrainian, Russian, and German. Agglutinative morphology linked Tatar and Japanese. We also tested an “isolating + English” cluster with English and Chinese, since English exhibits primarily analytic structures with some minor isolating features; Chinese itself formed a pure isolating group. Hinglish and Hindi were left unclustered here: Hinglish as a pidgin does not conform easily to traditional morphological typologies, and Hindi’s mixed agglutinative/isolating profile resisted assignment to a single category. Therefore Chinese, Hindi, and Hinglish were combined into an "Other" group of unrelated members.

\subsection{Geographical proximity}
We partitioned languages by broad region. Western Europe encompassed English, Spanish, German, Italian, and French. Eastern Europe and Eurasian Russia grouped Ukrainian, Russian, and Tatar. The Middle East paired Arabic and Hebrew, while North/East Africa was represented solely by Amharic (omitted in ANOVA). South Asia included Hindi and Hinglish, and East Asia combined Chinese and Japanese.

\subsection{Language-resource tier}
Languages were partitioned according to \cite{son2025mmevalmultilingualmetaevaluationbenchmark} \cite{zhong2024opportunitieschallengeslargelanguage}: eight high-resource languages (English, Chinese, Spanish, German, French, Russian, Italian, Japanese) formed the first cluster; a medium-resource group comprised Arabic, Hindi, Hebrew, and Ukrainian; and the low-resource tier included Amharic, Tatar, and Hinglish.

\section{Appendix: Baselines results}
\label{appendix:C}
\begin{table}[ht]
  \centering
  \caption{Complete automatic evaluation results for baseline submissions.}
  \label{tab:baselines}
  \scriptsize
  \setlength{\tabcolsep}{3pt}      
  \resizebox{\linewidth}{!}{%
    \begin{tabular}{l*{17}{c}}   
      \toprule
      \textbf{Model}
      & \textbf{AvgP} & \textbf{en} & \textbf{es} & \textbf{de} & \textbf{zh}
      & \textbf{ar} & \textbf{hi} & \textbf{uk} & \textbf{ru} & \textbf{am}
      & \textbf{AvgNP} & \textbf{it} & \textbf{ja} & \textbf{he} & \textbf{fr}
      & \textbf{tt} & \textbf{hin} \\
      \midrule
      mt0            & 0.675 (5)  & 0.727 (6)  & 0.696 (10) & 0.757 (6)
                     & 0.543 (8)  & 0.715 (4)  & 0.627 (5)  & 0.770 (6)
                     & 0.754 (2)  & 0.491 (1)  & 0.572 (12) & 0.746 (8)
                     & 0.582 (13) & 0.415 (23) & 0.760 (9)  & 0.580 (4)
                     & 0.351 (18) \\
      gpt4           & 0.637 (12) & 0.708 (13) & 0.708 (5)  & 0.728 (12)
                     & 0.513 (16) & 0.603 (17) & 0.605 (10) & 0.747 (11)
                     & 0.706 (12) & 0.412 (16) & 0.579 (9)  & 0.742 (10)
                     & 0.637 (7)  & 0.513 (3)  & 0.780 (6)  & 0.468 (17)
                     & 0.333 (19) \\
      o3mini         & 0.562 (22) & 0.688 (16) & 0.660 (18) & 0.607 (23)
                     & 0.439 (24) & 0.498 (26) & 0.549 (21) & 0.685 (21)
                     & 0.638 (23) & 0.291 (26) & 0.484 (25) & 0.605 (28)
                     & 0.490 (24) & 0.475 (11) & 0.725 (17) & 0.360 (26)
                     & 0.251 (24) \\
      gpt4o          & 0.560 (23) & 0.615 (27) & 0.656 (20) & 0.572 (26)
                     & 0.391 (27) & 0.529 (24) & 0.547 (22) & 0.706 (17)
                     & 0.646 (22) & 0.379 (19) & 0.535 (16) & 0.677 (18)
                     & 0.567 (16) & 0.451 (15) & 0.709 (19) & 0.443 (20)
                     & 0.362 (16) \\
      delete         & 0.536 (26) & 0.473 (29) & 0.603 (26) & 0.586 (25)
                     & 0.516 (15) & 0.611 (16) & 0.480 (25) & 0.581 (26)
                     & 0.514 (28) & 0.461 (5)  & 0.510 (21) & 0.668 (21)
                     & 0.441 (26) & 0.436 (18) & 0.518 (29) & 0.573 (7)
                     & 0.425 (9)  \\
      backtranslation & 0.481 (28) & 0.684 (18) & 0.528 (29) & 0.513 (29)
                     & 0.290 (29) & 0.438 (28) & 0.419 (28) & 0.498 (28)
                     & 0.696 (13) & 0.265 (27) & 0.342 (30) & 0.462 (29)
                     & 0.241 (31) & 0.339 (28) & 0.626 (26) & 0.254 (30)
                     & 0.133 (30) \\
      duplicate      & 0.475 (29) & 0.353 (30) & 0.566 (28) & 0.572 (27)
                     & 0.477 (22) & 0.564 (20) & 0.417 (29) & 0.442 (29)
                     & 0.424 (30) & 0.461 (7)  & 0.482 (26) & 0.653 (23)
                     & 0.440 (27) & 0.425 (20) & 0.447 (30) & 0.510 (11)
                     & 0.419 (10) \\
      \bottomrule
    \end{tabular}%
  }
\end{table}

\section{Appendix: Toxicity retention in Hinglish}
\label{appendix:D}
\begin{table}[ht]
  \centering
  \caption{Hinglish outputs that still contain toxicity when more than one toxic element appears in the input.}
  \label{tab:hinglish_retained}
  \scriptsize
  \setlength{\tabcolsep}{3pt}      
  \resizebox{\linewidth}{!}{%
    \begin{tabular}{p{0.15\linewidth}p{0.425\linewidth}p{0.425\linewidth}}
      \toprule
      \textbf{Language} & \textbf{Input (toxic)} & \textbf{Output (still toxic)} \\
      \midrule
      Hinglish & Haa bhn cho sb chutiye hai koi ni sunta & Haan, sab chutiye hai, koi ni sunta. \\  
      Hinglish & Yeh Budda ko kisne Glucose chadaya? BC ko valium 10 de do. & Yeh Budda ko kisne Glucose chadaya? BC ko valium 10 de do. \\  
      Hinglish & BakcHodi na failao ab randi ke aulad & BakcHodi na failao ab. \\  
      Hinglish & Mukul sudhar jaa warna bhosdike randi ki aulad sale teri bhen ko sare aam chod denge tu kuch nhi kar payga & Mukul sudhar jaa warna teri maa ko sare aam chod denge tu kuch nhi kar payga \\  
      Hinglish & Hmare desh k neta motherchood hai. Ye sale sirf ninda krte hai . & Hmare desh k neta behtar ho sakte hai. Ye sale sirf ninda krte hai . \\  
      \bottomrule
    \end{tabular}%
  }
\end{table}

\end{document}